\documentclass{article}
\usepackage{spconf,amsmath,graphicx}
\usepackage{enumitem}
\setlist{nosep, leftmargin=14pt}
\usepackage{mwe} 
\usepackage{url}

\usepackage{amsfonts}
\usepackage{amssymb}
\usepackage{tabularx}
\usepackage{multirow}
\usepackage{tabularx}
\usepackage[Export]{adjustbox}
\usepackage{caption}
\usepackage[skip=0pt]{caption}
\usepackage{graphicx}
\usepackage{subcaption}
\usepackage{subcaption}
\usepackage{xcolor}
\usepackage{multirow}
\usepackage{color, colortbl,booktabs}
\definecolor{Gray}{gray}{0.9}
\definecolor{myGreen}{rgb}{0.0, 0.6, 0.0}
\usepackage{pgfplots}
\pgfplotsset{compat=1.18}
\usepackage{sansmath}
\usepackage{todonotes}
\usepackage{pgfplots}
\usepackage{pgfplotstable}
\usepackage{marvosym}
\usepackage{pgfplots}
\pgfplotsset{compat=1.18}
\usepackage{hyperref}
\usepackage{float}

\makeatletter
\def\name#1{\gdef\@name{#1}}  
\makeatother

\title{Seeing Beyond the Image: ECG and Anatomical Knowledge-Guided Myocardial Scar Segmentation from Late Gadolinium-Enhanced Images}

\name{%
  Farheen Ramzan\textsuperscript{$\star$\Letter}\thanks{Correspondence: \textit{farheen.ramzan12@gmail.com};\textit{work.cherise@gmail.com}} \qquad
  Yusuf Kiberu\textsuperscript{$\ddagger$} \qquad
  Nikesh Jathanna\textsuperscript{$\ddagger$} \qquad
  Meryem Jabrane\textsuperscript{$\star$} \qquad \\
  Vicente Grau\textsuperscript{$\P$} \qquad
  Shahnaz Jamil-Copley\textsuperscript{$\ddagger\S$} \qquad
  Richard H.~Clayton\textsuperscript{$\star$} \qquad
  Chen (Cherise) Chen\textsuperscript{$\star$\Letter}
}

\address{
  $\star$ School of Computer Science, University of Sheffield, UK\\
  $\ddagger$ Department of Cardiology, Trent Cardiac Centre, Nottingham City Hospital, \\
  Nottingham University Hospitals NHS Trust, UK\\
  $\S$ School of Medicine, University of Nottingham, Nottingham, UK\\
  $\P$ Department of Engineering Science, University of Oxford, UK
}

\begin{document}

\maketitle

\begingroup
\renewcommand\thefootnote{}
\footnotetext{
\scriptsize \copyright\ 2026 IEEE. Personal use of this material is permitted. Permission from IEEE must be obtained for all other uses, in any current or future media, including reprinting/republishing this material for advertising or promotional purposes, creating new collective works, for resale or redistribution to servers or lists, or reuse of any copyrighted component of this work in other works.
}
\endgroup

\begin{abstract}
Accurate segmentation of myocardial scar from late gadolinium enhanced (LGE) cardiac MRI is essential for evaluating tissue viability, yet remains challenging due to variable contrast and imaging artifacts. Electrocardiogram (ECG) signals provide complementary physiological information, as conduction abnormalities can help localize or suggest scarred myocardial regions. In this work, we propose a novel multimodal framework that integrates ECG-derived electrophysiological information with anatomical priors from the AHA-17 atlas for physiologically consistent LGE-based scar segmentation. As ECGs and LGE-MRIs are not acquired simultaneously, we introduce a Temporal Aware Feature Fusion (TAFF) mechanism that dynamically weights and fuses features based on their acquisition time difference. Our method was evaluated on a clinical dataset and achieved substantial gains over the state-of-the-art image-only baseline (nnU-Net), increasing the average Dice score for scars from 0.6149 to 0.8463 and achieving high performance in both precision (0.9115) and sensitivity (0.9043). These results show that integrating physiological and anatomical knowledge allows the model to ``\textbf{see beyond the image}", setting a new direction for robust and physiologically grounded cardiac scar segmentation. Code is available at this \href{https://github.com/farheenjabeen/multimodal_ECG_LGE}{URL}.
\end{abstract}


\section{Introduction}

Late gadolinium enhancement (LGE) cardiac magnetic resonance imaging (CMR) is the gold standard for visualizing myocardial scar and fibrosis, providing critical insights into tissue viability and arrhythmic risk. Reliable segmentation of the left ventricular (LV) myocardium and scar regions from LGE MRI is essential for quantifying scar burden and guiding clinical decisions, yet automated segmentation of LGE-MRI remains challenging due to low signal-to-noise ratio, heterogeneous enhancement, motion artifacts, and scanner and pathology variability \cite{li2023myops}.

Deep neural networks such as U-Net \cite{ronneberger2015u} and nnU-Net \cite{isensee2021nnu} excel at medical image segmentation, yet purely image-based approaches struggle  with myocardial scar segmentation due to high contrast variability and imaging artifacts 
. We address this by integrating electrocardiography (ECG) and AHA-17 atlas to provide complementary electrophysiological context and shape prior. Although ECG is routinely used to localize scarred myocardial regions clinically \cite{bazoukis2022association}, its integration with cardiac MRI for pixel-level analysis remains largely underexplored. Existing multimodal scar segmentation frameworks \cite{qiu2023myops, li2023myops} mainly combine MRI sequences and are not directly applicable to ECG–MRI fusion. Joint learning from ECG and MRI is inherently challenging due to modality heterogeneity, dimensional differences, and temporal misalignment. In referral-based clinical settings such as the NHS in the UK, ECGs are often acquired weeks or months before MRI, and even simultaneous recordings are confounded by magnetohydrodynamic (MHD) artifacts \cite{li2024toward}.
\begin{figure*}[t]
    \centering
    \begin{subfigure}{0.99\columnwidth}
        \centering
        \includegraphics[width=0.85\linewidth]{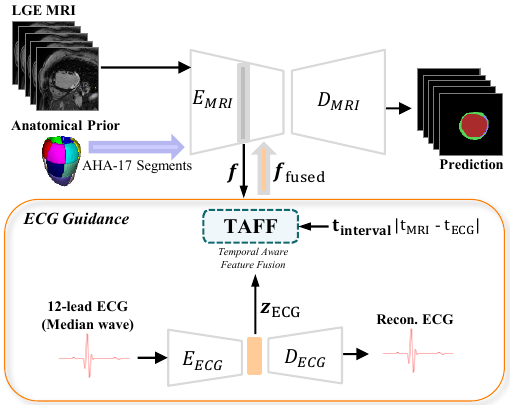}
        \caption{Our Multi-modal Network Architecture}
        \label{method1}
    \end{subfigure}
    \hfill
    \begin{subfigure}{0.99\columnwidth}
        \centering
        \includegraphics[width=0.65\linewidth]{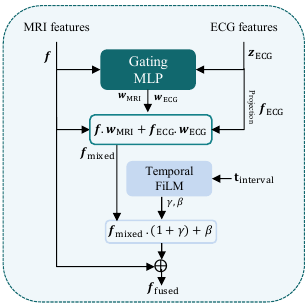}
        \caption{TAFF: Temporal Aware Feature Fusion}
        \label{method2}
    \end{subfigure}
    \caption{Overview of the proposed (a) multi-modal network combining LGE-MRI with AHA-based spatial prior, and ECG features via separate image and ECG autoencoders. (b) Temporal-Aware Feature Fusion (TAFF), where a gating network adaptively integrates ECG and MRI features, modulated by the ECG–MRI acquisition interval.}
    \label{model}
\vspace{-5pt}
\end{figure*}
To address these challenges, we develop a multimodal learning framework that jointly leverages electrical and anatomical information for improved myocardial scar segmentation from LGE images, featuring two key components: (1) a Temporal-Aware Feature Fusion (TAFF) module that adaptively integrates ECG and MRI features while accounting for their acquisition time difference, and (2) the 17-segment American Heart Association (AHA) model \cite{Cerqueira2002-AHA}, incorporated as an anatomical prior to provide spatial context and guide feature alignment. Together, these components enable more precise and physiologically consistent scar segmentation. Our main contributions are summarized as follows:
\begin{itemize}
    \item We design a novel Temporal-Aware Feature Fusion (TAFF) framework that accounts for the acquisition time difference between ECG and MRI. TAFF enables adaptive fusion of ECG and MRI features based on the informativeness and complementarity of each modality, while modulating their relative contributions according to the time interval between the two acquisitions.
    \item We introduce the 17-segment AHA model \cite{Cerqueira2002-AHA} as an anatomical prior to guide multimodal fusion and enhance the alignment between ECG-derived features and myocardial scar regions, significantly improving segmentation accuracy.
    \item To the best of our knowledge, this is \emph{one of the first studies} to harness ECG signals for guiding MRI-based myocardial scar segmentation, effectively bridging electrophysiological and structural domains to enable more physiologically grounded image analysis.
\end{itemize}

\section{Methods}
The proposed multi-modal network architecture for scar segmentation is shown in Fig.\ref{method1}, which features a novel Temporal Aware Feature Fusion (TAFF) (Fig.\ref{method2}) framework for ECG-MRI feature fusion, guided by a 17-segment AHA \cite{Cerqueira2002-AHA} anatomical prior for physiologically interpretable alignment and precise scar segmentation.

\subsection{Multi-modal Network Architecture}
At high-level, our network consists of an  image encoder $E_{MRI}$ which takes 2D MRI images and an AHA-17 map (see Sec.\ref{sec:atlas}) as input, followed by an image decoder $D_{MRI}$ for image segmentation; and an ECG autoencoder where an encoder $E_{ECG}$ is used to extract ECG features $\textbf{z}_{ECG} \in \mathbb{R}^{B \times 64}$ from 12-lead median waveforms and an ECG decoder $D_{ECG}$ is then applied for ECG signal reconstruction. To allow ECG-MRI feature fusion, we take the mid-level feature map $\textbf{f} \in \mathbb{R}^{B \times C \times H_{mid} \times W_{mid}}$ \footnote{In our experiments, we found that mid-level features outperform bottleneck features, offering a better balance between anatomical detail and semantic abstraction.} and fuse it with $\textbf{z}_{ECG}$ intelligently with the proposed TAFF module and send the fused feature back to the image branch through the remaining encoder and decoder stages to produce the final segmentation.

\subsection{TAFF: Temporal Aware Feature Fusion}\label{sec:TAFF}
As shown in Fig.~\ref{method2}, TAFF  adaptively integrates MRI and ECG features through a gating MLP that learns modality-specific weights ($w = [w_{MRI}, w_{ECG}] \in \mathbb{R}^{B \times 2}$, satisfying $\sum_i w_i = \textbf{1}$) to perform fusion between $\textbf{f}$ and the projected ECG features $\textbf{f}_{ECG} \in \mathbb{R}^{B \times C \times H_{mid} \times W_{mid}}$ from $\textbf{z}_{ECG}$: $\textbf{f}_{mixed} = \textbf{f}\times w_{MRI} + \textbf{f}_{ECG}\times w_{ECG}$. The mixed features are then modulated by a Temporal FiLM layer inspired by \cite{perez2018film}, which adjusts feature interactions based on the time interval between ECG and MRI acquisitions $t_{interval}=|t_{MRI}-t_{ECG}|$, producing temporally aligned and physiologically consistent fused features. Specifically, Temporal FiLM is a MLP,  which takes the $t_{interval}$ as input, and then produce affine scale $\gamma$ and shift parameters $\beta$ to calibrate features with time conditioning. We then apply residual connections to stabilize training which allows the network to neglect the feature fusion part when the ECG is not reliable and informative. The final fused feature is defined as: $\textbf{f}_{fused} = \textbf{f}+\textbf{f}_{mixed}\times (1+\gamma) + \beta$.

\subsection{AHA-17 Segment as Anatomical Prior}\label{sec:atlas}
To provide anatomically informed spatial guidance for multi-modal learning and segmentation, we generate a myocardial prior map based on the 17-segment AHA model \cite{Cerqueira2002-AHA}. Following established geometric implementations \cite{suinesiaputra2017statistical}, slice-level and angular partitioning are applied to 3D myocardium segmentation maps, which produces 17-channel binary masks. These 17-channel masks serve as spatial prior, which is concatenated directly with its corresponding MRI slice by slice to embed anatomical context, enabling region-aware feature learning and anatomically consistent scar segmentation.

\subsection{Loss Function}
The whole network is optimized with a composite loss: $\mathcal{L}_{total} = \mathcal{L}_{seg} + \lambda_{ECG}\mathcal{L}_{ecg} + \lambda_{ent}\mathcal{L}_{ent}$, where $\mathcal{L}_{seg}$ is the segmentation loss (consisting of Dice + cross-entropy) and $\mathcal{L}_{ecg}$ is the mean squared error loss between reconstructed and the true ECG signals. $\lambda_{ECG}$ is a coefficient, which gradually increases the ECG contribution during the first few warm-up epochs following a quadratic warmup schedule. We introduce a regularization loss based on the entropy of gating weights $w$ as $\mathcal{L}_{ent} = - \sum_i w_i log w_i$, which is used to prevent premature gating collapse during early stage of training \cite{chlon2025robust}. A small coefficient $\lambda_{ent}$ = 3e-3 keeps this regularization active without overpowering the primary loss components.

\vspace{-5pt}
\section{Experiments}
\noindent\textbf{Data.} We used a private Nottingham University Hospitals \cite{jathanna2023nottingham, shahnazprivatedata} dataset, comprising paired LGE-MRI scans and paper-based ECGs, keeping only cases where ECG and MRI were acquired within 90 days. ECGs were digitized from scanned papers using the open-source toolkit in \cite{fortune2022digitizing} with manual bounding boxes and visual quality control; discarding failed conversions. The LV blood pool, healthy myocardium, and scar were annotated by trained cardiologists with additional quality check. This yielded 103 high-quality MRI–ECG pairs, split 7:1:2 for training, validation, and testing. MRI data were preprocessed following \cite{isensee2021nnu} (resampling, normalization, and cropping), while median ECG waveforms were extracted using Hong et al. \cite{hong2017encase} to obtain standardized ECG representations.\\
\noindent\textbf{Implementation.} We employed the ECG dual-attention network \cite{Chen2025-ea} as the ECG autoencoder. The ECG-MRI time interval $t_{interval} = |t_{MRI}-t_{ECG}|$ (in days) was normalized to $[0,1]$. To stabilize multimodal training and prevent early gate collapse, each modality branch and the fusion module use separate optimizers. The MRI backbone was trained with stochastic gradient descent (SGD) (learning rate 1e-2 following a polynomial decay, momentum of 0.99, and weight decay 3e-5). The ECG network was first pretrained on the median waveforms of 12-lead ECGs from the PTB-XL dataset \cite{wagner2020ptb} (optimizer AdamW, wight decay 5e-5 with warmup and cosine decay) and then finetuned on our digitalized ECGs. The fusion gate parameters were optimized with AdamW as well (weight decay 1e-4) with a synchronized schedule for stable cross-modal adaptation.
\begin{table}[]
\centering
\caption{\textbf{Scar segmentation results on the test set: image-only baseline vs. proposed multimodal pipeline with module ablations.} Best scores in bold. \textit{(AP = Anatomical Prior)}. All improvements over the baseline are statistically significant (paired t-test, $p<0.001$).
}
\resizebox{\columnwidth}{!}{%
\begin{tabular}{@{}l|lc|rrrr@{}}
\toprule
\textbf{Model} & \textbf{Input Modality} & \textbf{w/ AP} & 
\textbf{Dice} & \textbf{Pre.} & \textbf{Sen} & \textbf{Vol. Diff.} \\ \midrule
Baseline (unimodal) & LGE & x & 0.6149 & 0.7315 & 0.7072 & 4.436 \\
\textbf{Ours (multimodal)} & \textbf{LGE+ECG} & \textbf{\checkmark} & \textbf{0.8463} & \textbf{0.9115} & \textbf{0.9043} & \textbf{1.971} \\ \midrule
\multirow{4}{*}{Ablation study} 
 & \begin{tabular}[c]{@{}l@{}}LGE+ECG\\ (w/o time conditioning)\end{tabular} & x & 0.5916 & 0.7351 & 0.6398 & 6.256 \\
 & \begin{tabular}[c]{@{}l@{}}LGE+ECG\\ (w/o time conditioning)\end{tabular} & \checkmark & 0.8234 & 0.8234 & 0.8900 & 2.039 \\
 & LGE & \checkmark & 0.8204 & 0.9090 & 0.8408 & 2.196 \\ \cmidrule(l){2-7}
 & \textbf{\begin{tabular}[c]{@{}l@{}}LGE+ECG\\ (w/ time conditioning)\end{tabular}} & \textbf{\checkmark} & \textbf{0.8463} & \textbf{0.9115} & \textbf{0.9043} & \textbf{1.971} \\ \bottomrule
\end{tabular}
}
\label{tab:results1}
\end{table}
\begin{figure*}[t]
    \centering
    \includegraphics[width=\textwidth]{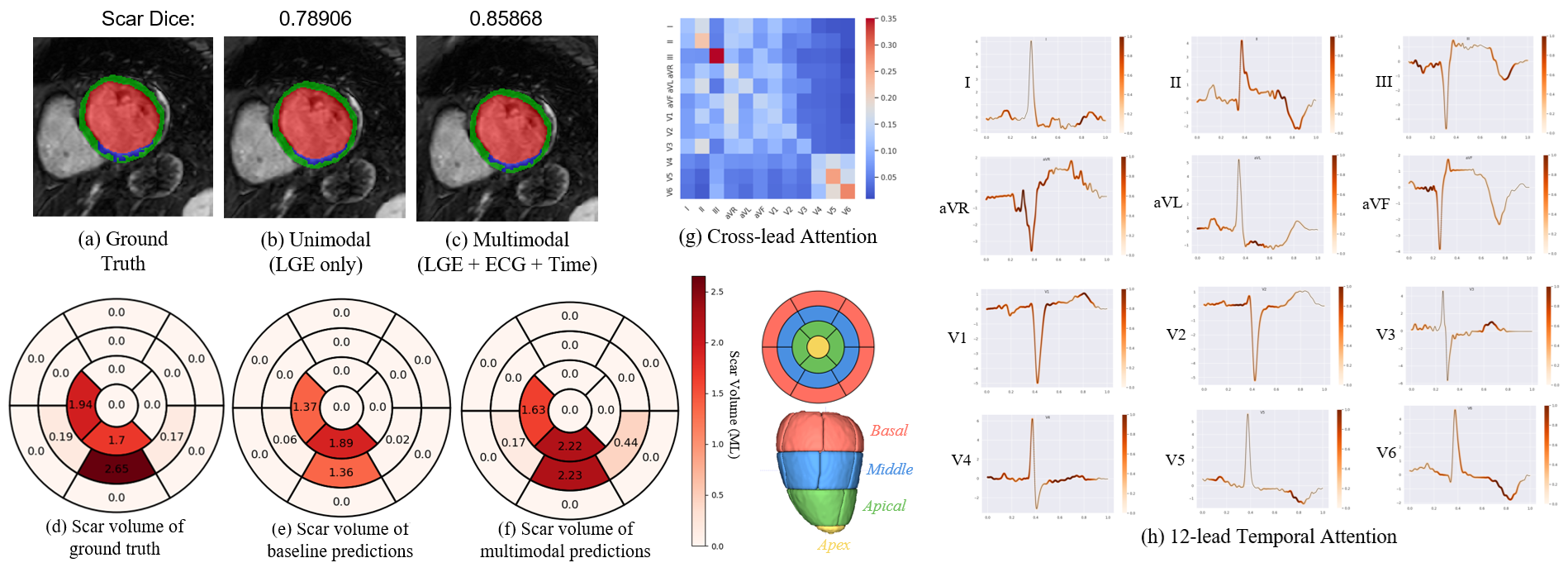}
    \caption{\textbf{Result Visualization:}  ground truth (a)  and segmentation predictions of \textcolor{red}{left ventricular blood pool}, \textcolor{myGreen}{healthy myocardium}, and \textcolor{blue}{scar} overlaid on a 2D LGE slice using unimodal (image-only) (b) and the proposed multimodal approach (c). We also plot the scar volume distribution on the AHA-17 segments accordingly (d-f); and cross-lead (g) and temporal lead-wise (h) attention maps for the corresponding ECG input using the ECG attention network to support model explainability.}
    \label{interpret}
\end{figure*}
\section{Results and Analysis}
Table \ref{tab:results1} presents test-set performance for our multi-modal pipeline versus the image-only baseline. We chose the same nnU-Net~\cite{isensee2021nnu} as the segmentation  backbone for fairness.\\
\noindent\textbf{Our multi-modal approach significantly outperforms the image-only baseline}. Using only LGE-MRI, the baseline achieves a Dice of 0.6149, reflecting the  challenges of scar segmentation under contrast variability and artifacts. Adding ECG features, anatomical priors, and time-aware fusion yields a markedly higher Dice score of 0.8463, demonstrating a substantial \emph{improvement over 37\%}. This demonstrates how physiological and anatomical information improves scar localization and boundary accuracy beyond image cues alone, as shown in Figure \ref{interpret}. Ablation results are  in Table \ref{tab:results1}. \\
\textbf{Directly fusing MRI and ECG is challenging}. Removing both the AHA-17 anatomical prior and the time-aware fusion (while the model still receives both LGE and ECG inputs) causes performance to drop to 0.5916, even lower than the image-only baseline (0.6149). This highlights the challenge of direct ECG-MRI fusion where the model struggles to find the spatial correspondence between electrical activity and myocardial regions.\\
\textbf{Contribution of anatomical prior.} Adding the anatomical prior significantly improves the Dice score from 0.5916 to 0.8234. This improvement can be attributed to two main factors: (i) the AHA-17 anatomical map acts as spatial regularization, constraining predictions within the myocardial wall and reducing anatomically implausible regions; and (ii) it provides fine-grained structural context that helps the model better understand myocardial anatomy and establish a good correspondence between ECG abnormalities and specific myocardial segments~\cite{Ortiz-Perez2008-rt}, thereby enhancing the physiological–anatomical alignment, with improved sensitivity from 0.8408 to 0.8900.\\
\textbf{Effectiveness of temporal conditioning.} Incorporating the acquisition time difference into the TAFF module further increases the Dice score from 0.8234 to 0.8463. This demonstrates that temporal conditioning enhances ECG–MRI alignment by adaptively modulating the influence of ECG features based on their recency, down-weighting outdated signals or less reliable signals and emphasizing temporally consistent information, leading to more coherent cross-modal alignment and higher segmentation accuracy.\\
\begin{figure}[]
\centering
\includegraphics[keepaspectratio=true, width=0.8\columnwidth]{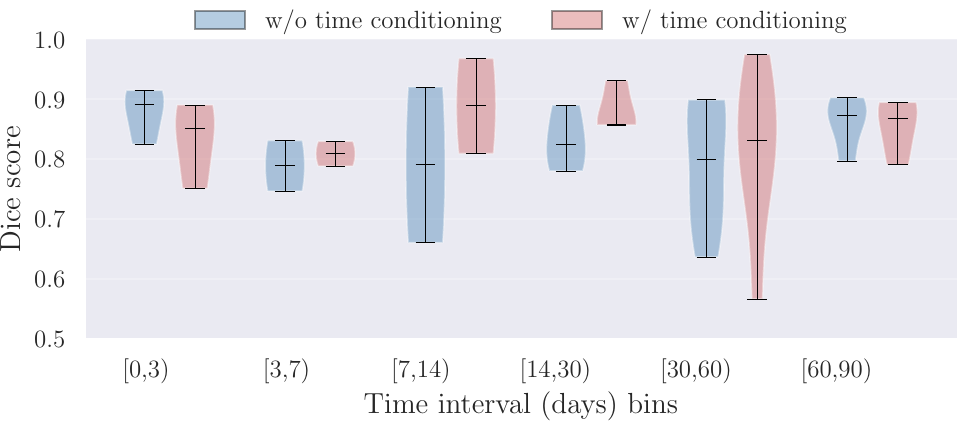}
\caption{Distribution of Dice scores across time-interval bins, comparing models with and without time conditioning.}\label{fig:dice_scores}
\end{figure}
\noindent\textbf{Result visualization and interpretability.} Figure \ref{interpret} compares segmentation quality (b,c) and region-wise scar volume across 17 AHA myocardial regions for the ground truth (d), the image-only baseline prediction (e), and our multimodal prediction (f) for a test subject. The ground truth plot shows scar burden in specific mid-ventricular and apical regions. The baseline model captures the broad pattern but consistently underestimates scar, while the multimodal model more closely matches both the magnitude and spatial distribution across the 17 myocardial regions. This reflects the added value of ECG and temporal cues in capturing subtle regional variations in scar distribution, resulting in uniform performance. The temporal attention maps (h) highlight that the model focuses more on the clinically indicative regions/biomarkers such as Q waves, QRS and/or abnormal ST–T segments~\cite{Das2008-oq,Kashou2025-dy}, on leads II, aVR, V2, and V4. The cross-lead attention map (g) highlights the distinctive pattern in Lead III and shows a positive correlation among V5, V6, and Lead II.\\
\noindent\textbf{Limitations and future work.} The ECG-driven gains are encouraging but remain modest, likely due to variability in digitalized ECG signal quality and the imbalanced sample distribution across bins. Future gains may be achieved by developing more robust digitization algorithms as well as designing better temporal-aware feature fusion with larger datasets and extending  2D to 3D. As shown in Fig.~\ref{fig:dice_scores}, the model with time conditioning achieved more significant improvements on cases with a 1-week [7,14) and 3-week to 1-month time window compared to the rest. The time-conditioned model shows higher median performance across most of the time bins, with more pronounced gains at longer time intervals indicating improved robustness to large temporal gaps between modalities. 
\vspace{-0.4cm}
\section{Conclusion}
This study shows that integrating ECG-derived electrophysiological information with anatomical priors from the AHA-17 atlas can substantially improve LGE-based scar segmentation. The proposed multimodal framework, supported by the Temporal Aware Feature Fusion mechanism, effectively handles the heterogeneity of the two modalities and their non-simultaneous acquisition. Evaluation on the clinical dataset demonstrated clear gains over the image-only baseline, with higher Dice, precision, and sensitivity. Overall, combining anatomical, functional, and temporal information resulted in more reliable cardiac scar segmentation.
\section{Compliance with ethical standards}
Ethical approval for clinical data collection was granted by the Health Research Authority (REC: 21/PR/0318; IRAS 290340; NUH reference 20CA015). The research conducted at Sheffield received ethical approval from the University Research Ethics Committee (reference: 064707; 01/10/2024).
\section{Acknowledgments}
F. R acknowledges support from a PhD scholarship funded by the Commonwealth Scholarship Commission in the United Kingdom. The project was supported by Royal Society (RGS/R2/242355). The data collection at Nottingham was supported via the MRC/NIHR CARP Grant (MR/V037595/1) led by Dr Jamil-Copley. Dr Jathanna and Dr Kiberu were supported by the Nottingham Hospitals Charity and Nottingham Research \& Innovations department.

\bibliographystyle{IEEEbib}
\bibliography{refs}

@ARTICLE{Chen2025-ea,
  title         = "Large Language Model-informed {ECG} Dual Attention Network
                   for Heart Failure Risk Prediction",
  author        = "Chen, Chen and others",
  year          =  2025,
  journal = "IEEE trans. on Big Data"
}

@article{jathanna2023nottingham,
  title={The Nottingham Ischaemic Cardiovascular Magnetic Resonance resource (NotIs CMR): a prospective paired clinical and imaging scar database—protocol},
  author={Jathanna, Nikesh and others},
  journal={Journal of Cardiovascular Magnetic Resonance},
  volume={25},
  number={1},
  pages={69},
  year={2023},
  publisher={Springer}
}

@MISC{shahnazprivatedata,
  title        = "Creation of a cardiac magnetic resonance ({CMR}) scar
                  segmentation tool using multi-vendor scans",
  booktitle    = "Health Research Authority",
  howpublished = "\url{https://www.hra.nhs.uk/planning-and-improving-research/}",
  note         = "Accessed: 2025-6-6",
  language     = "en"
}

@ARTICLE{Ortiz-Perez2008-rt,
  title     = "Correspondence between the 17-segment model and coronary arterial
               anatomy using contrast-enhanced cardiac magnetic resonance
               imaging",
  author    = "Ortiz-Pérez, José T and others",
  journal   = "JACC. Cardiovascular imaging",
  publisher = "Elsevier BV",
  volume    =  1,
  number    =  3,
  pages     = "282--293",
  month     =  may,
  year      =  2008,
  language  = "en"
}

@article{isensee2021nnu,
  title={nnU-Net: a self-configuring method for deep learning-based biomedical image segmentation},
  author={Isensee, Fabian and others},
  journal={Nature methods},
  volume={18},
  number={2},
  pages={203--211},
  year={2021},
  publisher={Nature Publishing Group}
}

@article{Cerqueira2002-AHA,
  title={Standardized myocardial segmentation and nomenclature for tomographic imaging of the heart: a statement for healthcare professionals from the Cardiac Imaging Committee of the Council on Clinical Cardiology of the American Heart Association},
  author={Cerqueira, MD and others},
  journal={Circulation},
  volume={105},
  number={4},
  pages={539--542},
  year={2002}
}

@inproceedings{ronneberger2015u,
  title={U-net: Convolutional networks for biomedical image segmentation},
  author={Ronneberger, Olaf and others},
  booktitle={International Conference on Medical image computing and computer-assisted intervention},
  pages={234--241},
  year={2015},
  organization={Springer}
}

@article{bazoukis2022association,
  title={Association of electrocardiographic markers with myocardial fibrosis as assessed by cardiac magnetic resonance in different clinical settings},
  author={Bazoukis, George and others},
  journal={World Journal of Cardiology},
  volume={14},
  number={9},
  pages={483},
  year={2022}
}

@ARTICLE{Das2008-oq,
  title     = "Fragmented wide {QRS} on a 12-lead {ECG}: a sign of myocardial
               scar and poor prognosis: A sign of myocardial scar and poor
               prognosis",
  author    = "Das, Mithilesh K and others",
  journal   = "Circulation. Arrhythmia and electrophysiology",
  publisher = "Ovid Technologies (Wolters Kluwer Health)",
  volume    =  1,
  number    =  4,
  pages     = "258--268",
  month     =  oct,
  year      =  2008,
  language  = "en"
}

@INCOLLECTION{Kashou2025-dy,
  title     = "{ST} segment",
  author    = "Kashou, Anthony H and others",
  booktitle = "StatPearls",
  publisher = "StatPearls Publishing",
  address   = "Treasure Island (FL)",
  month     =  jan,
  year      =  2025,
  language  = "en"
}

@inproceedings{hong2017encase,
  title={ENCASE: An ENsemble ClASsifiEr for ECG classification using expert features and deep neural networks},
  author={Hong, Shenda and others},
  booktitle={2017 Computing in cardiology (cinc)},
  pages={1--4},
  year={2017},
  organization={IEEE}
}

@article{fortune2022digitizing,
  title={Digitizing ECG image: A new method and open-source software code},
  author={Fortune, Julian D and others},
  journal={Computer methods and programs in biomedicine},
  volume={221},
  pages={106890},
  year={2022},
  publisher={Elsevier}
}

@article{li2024toward,
  title={Toward enabling cardiac digital twins of myocardial infarction using deep computational models for inverse inference},
  author={Li, Lei and others},
  journal={IEEE transactions on medical imaging},
  volume={43},
  number={7},
  pages={2466--2478},
  year={2024},
  publisher={IEEE}
}

@article{qiu2023myops,
  title={MyoPS-Net: Myocardial pathology segmentation with flexible combination of multi-sequence CMR images},
  author={Qiu, Junyi and others},
  journal={Medical image analysis},
  volume={84},
  pages={102694},
  year={2023},
  publisher={Elsevier}
}

@article{li2023myops,
  title={MyoPS: A benchmark of myocardial pathology segmentation combining three-sequence cardiac magnetic resonance images},
  author={Li, Lei and others},
  journal={Medical Image Analysis},
  volume={87},
  pages={102808},
  year={2023},
  publisher={Elsevier}
}

@article{chlon2025robust,
  title={Robust Multimodal Learning via Entropy-Gated Contrastive Fusion},
  author={Chlon, Leon and others},
  journal={arXiv preprint arXiv:2505.15417},
  year={2025}
}

@inproceedings{perez2018film,
  title={Film: Visual reasoning with a general conditioning layer},
  author={Perez, Ethan and others},
  booktitle={Proceedings of the AAAI conference on artificial intelligence},
  volume={32},
  number={1},
  year={2018}
}

@article{suinesiaputra2017statistical,
  title={Statistical shape modeling of the left ventricle: myocardial infarct classification challenge},
  author={Suinesiaputra, Avan and others},
  journal={IEEE journal of biomedical and health informatics},
  volume={22},
  number={2},
  pages={503--515},
  year={2017},
  publisher={IEEE}
}

@article{wagner2020ptb,
  title={PTB-XL, a large publicly available electrocardiography dataset},
  author={Wagner, Patrick and others},
  journal={Scientific data},
  volume={7},
  number={1},
  pages={1--15},
  year={2020},
  publisher={Nature Publishing Group}
}

\end{document}